\documentclass[final]{cvpr}

\usepackage{times}
\usepackage{epsfig}
\usepackage{graphicx}
\usepackage{amsmath}
\usepackage{amssymb}
\usepackage{acronym}

\usepackage{multirow}
\usepackage{multicol}
\usepackage{graphicx}
\usepackage{verbatim}
\usepackage{float}

\usepackage{ragged2e}
\usepackage{subcaption}
\usepackage[pagebackref=true,breaklinks=true,colorlinks,bookmarks=false]{hyperref}

\def\cvprPaperID{10109} 
\def\confYear{CVPR 2021}

\begin{document}

\title{Multiresolution Knowledge Distillation for Anomaly Detection}

\author{Mohammadreza Salehi, Niousha Sadjadi*, Soroosh Baselizadeh*, Mohammad Hossein Rohban, Hamid R. Rabiee\\

Sharif University of Technology\\
{\tt\small (smrsalehi, nsadjadi, baselizadeh)@ce.sharif.edu, 
(rohban, rabiee)@sharif.edu}
}



\maketitle

\begin{abstract}


Unsupervised representation learning has proved to be a critical component of anomaly detection/localization in images.
The challenges to learn such a representation are two-fold. Firstly, the sample size is not often large enough to learn a rich generalizable representation through conventional techniques. Secondly, while only normal samples are available at training, the learned features should be discriminative of normal and anomalous samples. Here, we propose to use the ``distillation'' of features at various layers of an expert network, pre-trained on ImageNet, into a simpler cloner network to tackle both issues. 
We detect and localize anomalies using the discrepancy between the expert and cloner networks' intermediate activation values given the input data.
We show that considering multiple intermediate hints in distillation leads to better exploiting the expert's knowledge and more distinctive discrepancy compared to solely utilizing the last layer activation values.
Notably, previous methods either fail in precise anomaly localization or need expensive region-based training. In contrast, with no need for any special or intensive training procedure, we incorporate interpretability algorithms in our novel framework for localization of anomalous regions. Despite the striking contrast between some test datasets and ImageNet, we achieve competitive or significantly superior results compared to the SOTA methods on MNIST, F-MNIST, CIFAR-10, MVTecAD, Retinal-OCT, and two Medical datasets on both anomaly detection and localization.
\end{abstract}
\section{Introduction}

\begin{figure}[!t]
     \label{Abstract}
    \includegraphics[width=\linewidth]{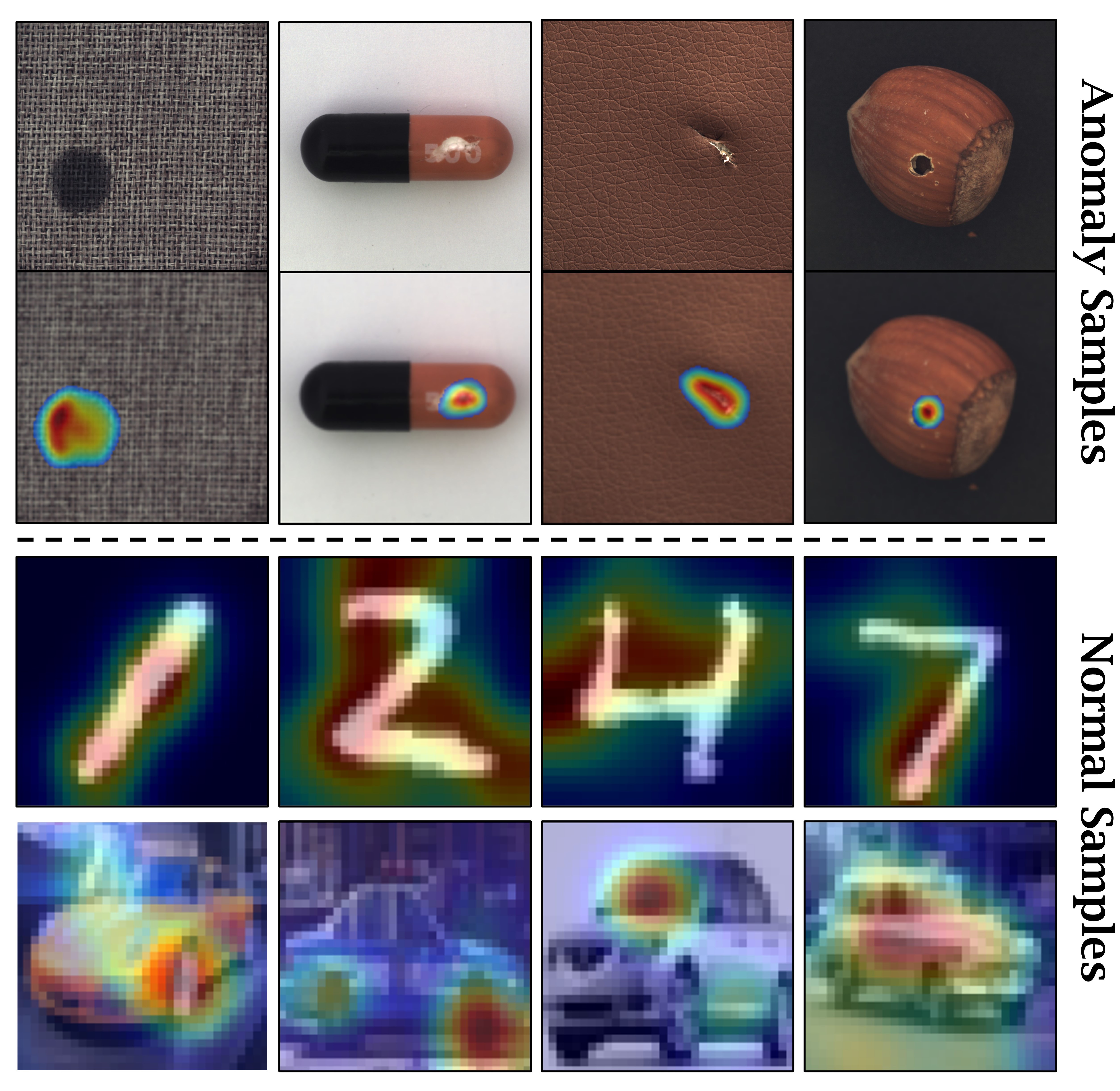}
    \caption{Our precise heatmaps localizing anomalous features in MVTecAD (top two rows) and normal features in MNIST and CIFAR-10 (two bottom rows).}
\end{figure}
Anomaly detection (AD)
aims for recognizing test-time inputs looking abnormal or novel to the model according to the previously seen normal samples during training. It has been a vital demanding task in computer vision with various applications, like in industrial image-based product quality control~\cite{mei2018automatic, bergmann2019mvtec} or in health monitoring processes~\cite{li2018thoracic}. These tasks also require the pixel-precise localization of the anomalous regions, called defects. This is pivotal for comprehending the dynamics of monitored procedures and triggering the apt antidotes, and providing proper data for the downstream models in industrial settings.

Traditionally, the AD problem has been approached in a one-class setting, where the anomalies represent a broadly different class from the normal samples. Recently, considering subtle anomalies has attracted attentions. This new setting further necessitates precise anomaly localization. However, performing excellently in both settings on various datasets is highly appreciated but is not yet fully achieved.

Due to the unsupervised nature of the AD problem and the restricted data access, availability of just the normal data in training, the majority of methods \cite{sabokrou2018adversarially, perera2019ocgan, schlegl2019f, golan2018deep, ruff2019deep} model the normal data abstraction by extracting semantically meaningful latent features. 
These methods perform well solely on either of the two mentioned cases. This problem, called the \textit{generality} problem~\cite{salehi2020puzzle}, highly declines trust in them on unseen future datasets.
Moreover, anomaly localization is either impossible or poor in most of them \cite{sabokrou2018adversarially, perera2019ocgan, ruff2018deep} and leads to intensive computations that hurt their real-time performance. Additionally, many earlier works \cite{ruff2018deep, perera2019ocgan} suffer from unstable training, requiring unprincipled early stopping to achieve acceptable results. 


Using the pre-trained networks, though not fully explored in the AD context, could potentially be an alternative track. This is especially very helpful when the sample size is small and the normal class shows large  variations. Some earlier studies \cite{andrews2016transfer, burlina2019s, Napoletano2018cnn-based, nazare2018pre} try to train their model based on the pre-trained features of normal data. These methods either miss anomaly localization \cite{andrews2016transfer, burlina2019s}, or tackle the problem in a region-based fashion \cite{Napoletano2018cnn-based, zhang2013region}, i.e. splitting images into smaller patches to determine the sub-regional abnormality. This is computationally expensive and often leads to inaccurate localization. 
To evade this issue, Bergmann \etal \cite{bergmann2020uninformed} train an ensemble of student networks to mimic the \textit{last layer} of a teacher network on the anomaly-free data. However, performing a region-based approach in this work, not only makes it heavily rely on the size of the cropped patches and hence susceptible to the changes in this size, but also intensifies the training cost severely. Furthermore, imitating only the last layer misses to fully exploit the knowledge of the teacher network \cite{romero2014fitnets}. This makes them complicate their model and employ other complementary techniques, such as self-supervised learning, in parallel. 

Lately, Zhang \etal \cite{zhang2018unreasonable} have demonstrated that the activation values of the intermediate layers of neural networks are a firm perceptual representation of the input images. 
By this premise, we propose a novel knowledge distillation method that is designed to \textit{distill} the \textit{comprehensive} knowledge of an ImageNet pre-trained \textit{source} network, \textit{solely} on the normal training data, into a simpler \textit{cloner} network. 
This happens by forcing the cloner's \textit{intermediate} embedding of normal training data at \textit{several critical layers} to conform to those of the source.
Consequently, the cloner learns the manifold of the normal data thoroughly, and yet earns no knowledge from the source about other possible input data. Hence, the cloner will behave differently from the source when fed with anomalous data. Furthermore, a simpler cloner architecture enables avoiding distraction by non-distinguishing features, and  enhances the discrepancy in behavior of the two networks on anomalies.

In addition, we derive precise anomaly localization heat maps, without using region-based expensive training and testing, through exploiting the concept of gradient. We evaluate our method on a comprehensive set of datasets on various tasks of anomaly detection/localization where we exceed the SOTA in both localization and detection. Our training is highly stable and needs no dataset-dependent fine tuning. As we only train the cloner's parameters, we require just \textit{one} more \textit{forward} pass of inputs through the source compared to a standard network training on the normal data. We also investigate our method through exhaustive ablation studies. Our main contributions are summarized as follows:



\begin{enumerate}
    
    \item Enabling a more comprehensive transfer of the knowledge of the pre-trained expert network to the cloner one. Distilling the knowledge into a \textit{more compact} network also helps concentrating solely on the features that are distinguishing normal vs. anomalous.
    \item Our method has a computationally inexpensive and stable training process compared to the earlier work.
    
    \item Our method allows a real-time and precise  anomaly localization based on computing gradients of the discrepancy loss with respect to the input.
  
    \item Conducting a huge number of diverse experiments, and outperforming previous SOTA models by a \textit{large} margin on many datasets and yet staying competitive on the rest.

\end{enumerate}

\begin{figure*}[!t]
\centering
     \includegraphics[width=\textwidth]{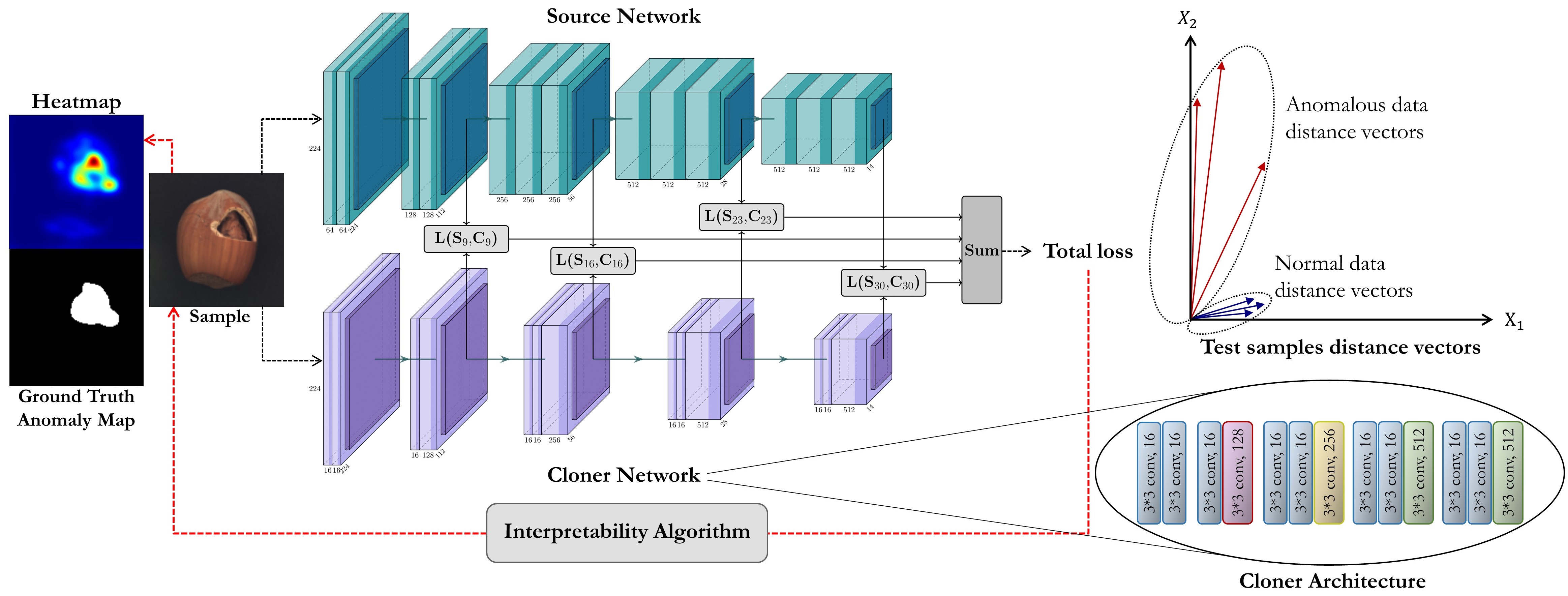}
      \caption{Visualized summary of our proposed framework. A smaller cloner network, $C$, is trained to imitate the \textit{whole} behavior of a source network, $S$, (VGG-16) on normal data. The discrepancy of their intermediate behavior is formulated by a total loss function and used to detect anomalies test time. A hypothetical example of distance vectors between the activations of $C$ and $S$ on anomalous and normal data is also depicted. Interpretability algorithms are employed to yield pixel-precise anomaly localization maps. }
       \label{network visualization}
\end{figure*}
\section{Related Work}
\label{sec:related_work}

\textbf{Previous Methods:}
Autoencoder(AE)-based methods use the idea that by learning normal latent features, abnormal inputs are not reconstructed as precise as the normal ones. This results in higher reconstruction error for anomalies. 
To better learn these normal latent features, LSA~\cite{abati2019latent} trains an autoregressive model at its latent space and OC-GAN~\cite{perera2019ocgan} attempts to force abnormal inputs to be reconstructed as normal ones. These methods fail on industrial or complex datasets~\cite{salehi2020arae}. SSIM-AE~\cite{bergmann2018improving} trains an AE with SSIM loss~\cite{zhao2015loss} instead of MSE causing to perform just better on defect segmentation. Gradient-based VAE~\cite{Dehaene2020} introduces an energy criterion, which is minimized at test-time by an iterative procedure. Both of the mentioned methods do not perform well on one-class settings, such as CIFAR-10~\cite{cifar}.



GAN-based approaches, like AnoGan~\cite{schlegl2017unsupervised}, f-AnoGan~\cite{schlegl2019f}, and GANomaly~\cite{akcay2018ganomaly}, attempt to find a specific latent space where the generator's reconstructions, obtained from samplings of this space, are analogous to the normal data. f-AnoGan and GANomaly add an extra encoder to the generator to reduce inference time of AnoGan. Despite their acceptable performance in localization and detection on subtle anomalies, they fail on one-class settings.

Methods like uninformed-students~\cite{Bergmann2020ustd}, GT\cite{golan2018deep}, and DSVDD~\cite{ruff2018deep} keep only the useful information of normal data by building a compact latent feature space, in contrast to AE-based ones that try to miss the least amount of normal data information. To achieve this, they use self-supervised learning methods or one-class techniques. However, since we only have access to normal samples in an unsupervised setting, the optimization here is harder than in AE-based methods and usually converges to trivial solutions. To solve this issue, unprincipled early stopping is used that lowers the trust in these models on unseen future datasets. For example, GT fails on subtle anomaly datasets like MVTecAD while performs well on one-class settings.


\textbf{Using Pre-trained Features:} Some previous methods use pre-trained VGG's last layer to solve the representation problem \cite{chen2001one, sabokrou2018deep}. However, \cite{chen2001one} sticks in bad local minima as it uses only the last layer. \cite{sabokrou2018deep} attempts to solve this by extracting lots of different patches from normal images. Then, it fits a Gaussian distribution on the VGG extracted embeddings of the patches. Although this might alleviate the problem, they fail to provide good localization or detection on diverse datasets because of using unimodal Guassian distribution and hand engineered size of patches. 



\textbf{Interpretability Methods:} Determining the contribution of input elements to a deep function is investigated in interpretability methods. Gradient-based methods computes pixel's importance using gradients as a proxy. While Gradients~\cite{simonyan2013deep} uses rough gradients, GuidedBackprop (GBP)~\cite{springenberg2014striving} filters out negative backpropagated gradients to only consider elements with positive contribution. As Gradients' maps can be noisy, SmoothGrad~\cite{smilkov2017smoothgrad} adds small noises to the input and averages the maps obtained using Gradients for each noisy input. Several methods \cite{adebayo2018sanity, nie2018theoretical} reveal some flaws in GBP by demonstrating that it reconstructs the image instead of explaining the outcome function.






\section{Method}
\label{method}


\subsection{Our Approach}
Given a training dataset $D_{train}=\lbrace x_1, ..., x_n \rbrace$ consisting only of normal images (i.e. no anomalies in them), we ultimately train a \textit{cloner} network, $C$, that detects anomalous images in the test set, $D_{test}$, and localizes anomalies in those images with the help of a pre-trained network. As $C$ needs to predict the deviation of each sample from the manifold of normal data, it needs to know the manifold quite well.
Therefore, it is trained to mimic the \textit{comprehensive} behavior of an expert network, called the \textit{source} network $S$. Earlier Work in knowledge distillation have conducted huge efforts to transfer one network's knowledge to another smaller one for saving computational cost and memory usage. Many of them strive to teach just the output of $S$ to $C$. We, however, aim to transfer the intermediate knowledge of $S$ on the normal training data to $C$ as well.

In~\cite{romero2014fitnets}, it is shown that by using a single intermediate level hint from the source, thinner but deeper cloner even outperforms the source on classification tasks. In this work, we provide $C$ with multiple intermediate hints from $S$ by encouraging $C$ to learn $S$'s knowledge on normal samples through conforming its intermediate representations in a number of \textit{critical layers} to $S$'s representations. It is known that layers of neural networks correspond to features at various abstraction levels. For instance, first layer filters act as simple edge detectors. They represent more semantic features when considering later layers. Therefore, mimicking different  layers, educates $C$ in various abstraction levels, which leads to a more thorough final understanding of normal data. In contrast, using only the final layer shares a little portion of $S$'s knowledge with $C$. In addition, this causes the optimization to stuck in irrelevant local minima. On the contrary, using several intermediate hints turns the ill-posed problem into a more well-posed one. The effect of considering different layers is more investigated in Sec. \ref{ablation:layers}.


In what follows, we refer to the $i$-th critical layer in the networks as $CP_i$ ($CP_0$ stands for the raw input) and the source activation values of that critical layer as $a_{s}^{CP_i}$, and the cloner's ones as $a_{c}^{CP_i}$. As discussed in knowledge distillation literature \cite{romero2014fitnets, yim2017gift}, the notion of knowledge can be seen as the value of activation functions. We define the notion of knowledge as both the value and direction of all $a^{CP_i}$s to intensify the full knowledge transfer from $S$ to $C$. Hence, we define two losses, $\mathcal{L}_{val}$ and $\mathcal{L}_{dir}$ to represent each aspect. The first, $\mathcal{L}_{val}$, aims to minimize the Euclidean distance between $C$'s and $S$'s activation values at each $CP_i$. Thus, $\mathcal{L}_{val}$ is formulated as
\begin{equation}
\label{train_eq1}
\resizebox{.8 \linewidth}{!}{$
\mathcal{L}_{val}=\sum_{i=1}^{N_{CP}}\frac{1}{N_i}\sum_{j=1}^{N_i}(a_s^{CP_i}(j) - a_c^{CP_i}(j))^2$},
\end{equation}

where $N_i$ indicates the number of neurons in layer $CP_i$ and $a_{.}^{CP_i}(j)$ is the value of $j$-th activation in layer $CP_i$. $N_{CP}$ represents total number of critical layers.

Additionally, we use the $\mathcal{L}_{dir}$ to increase the directional similarity between the activation vectors. This is more vital in ReLU networks whose neurons are activated only after exceeding a zero value threshold. This indicates that two activation vectors with the same Euclidean distance from the target vector, may have contrasting behaviors in activating a following neuron. For instance, for $e$ being a positive number, let $a_1 = (0, 0, e, 0, \dots, 0) \in \mathbb{R}^{k}, a_2 = (0, (\sqrt{2}+1)e, 0, 0, \dots, 0) \in \mathbb{R}^{k}$
be activation vectors of two disparate cloner networks both trying to mimic the activation vector of a source network, $a^{*}$, defined as $a^{*} = (0, e, 0, 0, \dots, 0) \in \mathbb{R}^{k}$.
It is clear $a_1$ and $a_2$ have the same Euclidean distance from $a^{*}$. However, assuming $W=(0, 1, \dots, 0, 0)$ as weight vector of a neuron in the next layer of the network, we have 
\begin{equation}
    \begin{split}
      W^{T}a_1 &= 0 \leq 0, \\
      W^{T}a_2 &= (\sqrt{2} + 1)e > 0,\\
      W^{T}a^{*} &= e > 0.
    \end{split}
\end{equation} 
This means that the corresponding ReLU neuron would be activated by $a_2$, similar to $a^*$, while deactivated by $a_1$. To address this, using the cosine similarity metric, we define the $\mathcal{L}_{dir}$ as

\begin{equation}
\label{train_eq2}
\mathcal{L}_{dir} = 1-\sum_i \frac{vec(a_s^{CP_i})^T \cdot vec(a_c^{CP_i})}{\left \| vec(a_s^{CP_i}) \right \| \left \| vec(a_c^{CP_i}) \right \|},
\end{equation}

where $vec(x)$ is a vectorization function transforming a matrix $x$ with arbitrary dimensions into a 1-D vector. This encourages the activation vector of $C$ be not only close to the $S$'s one in terms of Euclidean distance but also be in the same direction. Note that $\mathcal{L}_{dir}$ is $1$ for $a_1$, and is $0$ for  $a_2$. The role of $\mathcal{L}_{dir}$ and $\mathcal{L}_{val}$ is more elaborated in Sec. \ref{ablation:dir_loss}. Using the two aforementioned losses, $\mathcal{L}_{total}$ is formulated as 
\begin{equation}
\label{total_eq}
\mathcal{L}_{total} = \mathcal{L}_{val} + \lambda \mathcal{L}_{dir},
\end{equation}
where $\lambda$ is set to make the scale of both constituent terms the same. For this, we find the initial amount of error for each term on the untrained network and set $\lambda$ with respect to it. Training using $\mathcal{L}_{total}$, unlike many other methods \cite{golan2018deep, bergman2020classification}, continues to fully converge, which is the only accessible criterion to measure when to stop training epochs.

Moreover, the architecture of $C$ is designed to be simpler than $S$ to enable knowledge ``distillation". This compression of the network facilitates the concentration on normal main features. While the source needs to be a very deep wide model to learn all necessary features to perform well on a large-scale domain dataset, like ImageNet \cite{deng2009imagenet}, the goal of the cloner is simply acquiring the source's knowledge of the normal data. Hence, superfluous filters are only detrimental by focusing on non-distinguishing features, present in both normal and anomalous data. Compressing the source prevents such distractions for the model. This can be of a greater vitality when dealing with normal data having a more restricted scope. The effect of the cloner's architecture is explored in Sec. \ref{ablation:compression}.

\textbf{Anomaly Detection:} To detect anomalous samples, each test input is fed to both $S$ and $C$. As $S$ has only taught the normal point of view to $C$, anomalies, inputs out of the normal manifold, are a potential surprise for $C$. On the other hand, $S$ is knowledgeable on anomalous inputs too. All this leads to a potential discrepancy in their behavior with anomalous inputs that is thresholded for anomaly detection using Eq. \ref{total_eq}, which formulates this discrepancy.

\textbf{Anomaly Localization:}  \cite{Dehaene2020, zimmerer2019case} have  shown that the derivative of loss function with respect to the input has meaningful information about the significance of each pixel. We employ gradients of $\mathcal{L}_{total}$ to find anomalous regions causing an increase in its value. To obtain our localization map for the input $x$, we first acquire the attribution map, $\Lambda$ by
\begin{equation}
\label{derivation_eq}
\Lambda = \frac{\partial \mathcal{L}_{total} }{\partial x}.
\end{equation}
To reduce the natural noises in these maps, we induce Gaussian blur and opening morphological filter on $\Lambda$. Hence, the localization map, $L_{map}$, is achieved by
\begin{equation}
\label{local_eq}
    \begin{split}
        M &= g_{\sigma}(\Lambda),\\
        L_{map} &= (M \ominus B) \oplus B,
    \end{split}
\end{equation}
where g denotes a Gaussian filter with standard deviation of $\sigma$. $\ominus$ and $\oplus$ represent morphological erosion and dilation by a structuring element $B$, respectively. Together, called opening, these operations remove small sporadic noises and yield clean maps. The structuring element, $B$, is a simple binary map usually in shape of an ellipse or disk. Instead of using simple gradients as in Eq. \ref{derivation_eq}, some other gradient-based interpretability methods can be employed to further illuminate the role of each pixel on loss value. We discuss different methods more in Sec. \ref{ablation:localization}.
Our proposed framework is shown schematically in Figure \ref{network visualization}. Note that we need only two forward passes for detection and one backward pass through $C$ for localization.

\subsection{Settings}
VGG~\cite{simonyan2014very} features have shown great performance in classification and transfer learning \cite{tan2018survey, weiss2016survey}. This highlights the practicality of its filters in different domains. By transferring the knowledge of an ImageNet VGG-16 to a simple cloner, we exploit the discrepancy of features between $C$ and $S$ to find anomalies. In our VGG-16 source network, we choose the four final layers of each convolutional block, i.e. max-pooling layers, to be the critical points ($CP_i$s). Selecting critical points is  explored more in Sec. \ref{ablation:layers}. 

For the cloner network, for all experiments and datasets, we use the architecture described in Figure \ref{network visualization}, which is smaller than the source. As a result, it can benefit from the advantages of compression discussed in Sec. \ref{method}. The role of cloner architecture is discussed more in Sec. \ref{ablation:compression}. Note that, similar to \cite{ruff2018deep}, we avoid using bias terms in our cloner's network. As proven by \cite{ruff2018deep}, networks with bias in any layer can easily learn constant functions, independent of the input. In our work, though it can be negligible on datasets with diverse normal data, it can be detrimental when normal images are roughly the same. To be more specific, for some layers $l$ and $l+1$ that are between any $i$-th and $(i-1)$-th $CP$, the cloner can generate a specific constant activation vector, $a_C^{CP_i}$, regardless of the input, only by setting the $l$-th layer's weight to zero and adjusting the $l+1$-th layers's bias. As the normal training images are much alike, the source's intermediate activations are also highly similar for them. Therefore, those constant $a_c^{CP_i}$s can be arbitrarily close to the source's correlated intermediate activations for any training input, which is the goal of training phase while harming the test procedure since they are constant outputs indeed. To avoid this, we use a bias-less network for $C$.

In all experiments, we use Adam optimizer~\cite{kingma2014adam} with learning rate $=0.001$ and batch size $=64$ for optimization.

\subsection{Ablation Studies}
\subsubsection{Intermediate Knowledge}
\label{ablation:layers}
In this experiment, we examine the effect of involving the last, the last two, and the last four max-pooling layers as $CP_i$s on MVTecAD and MNIST. We report average AUROC of all classes in Figure \ref{ablation:layers}. Obviously, a consistent growing trend exist that shows the effectiveness of considering more layers. Notice that some MVTecAD classes (e.g ``screw") have near random AUCROC in ``just the last layer`` setting. This suggests that using just the last layer makes the problem ill-posed and hard to optimize.


\begin{figure}[h]
\centering
     \includegraphics[width=\linewidth]{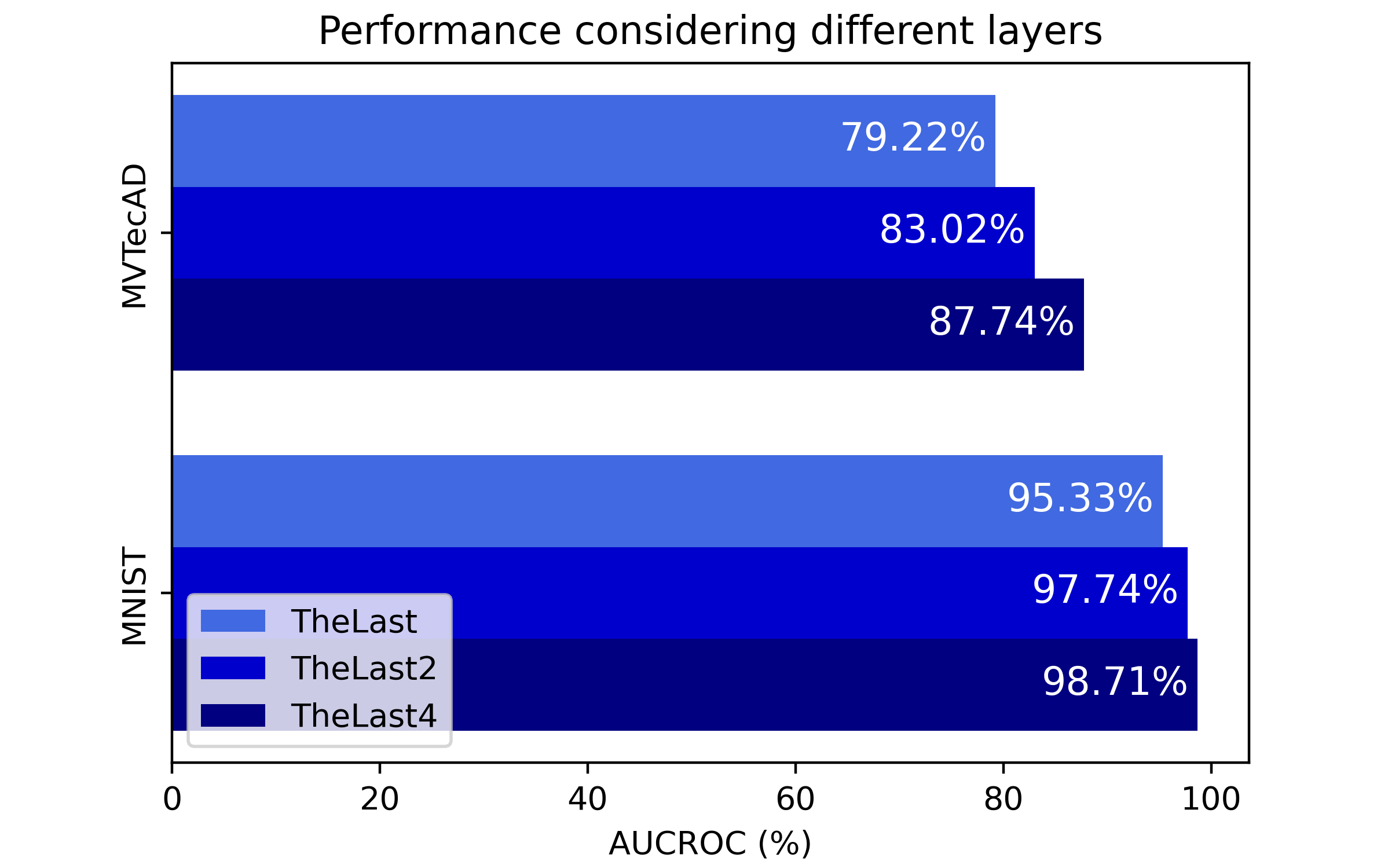}
      \caption{The performance of our proposed method using various layers for distillation. More intermediate layers lead to a performance boost on anomaly detection.}
       \label{fig:ablation_layers}
\end{figure}

\subsubsection{Distillation Effect (Compact $C$)}
\label{ablation:compression}
\begin{figure}[!h]
\centering
     \includegraphics[width=\linewidth]{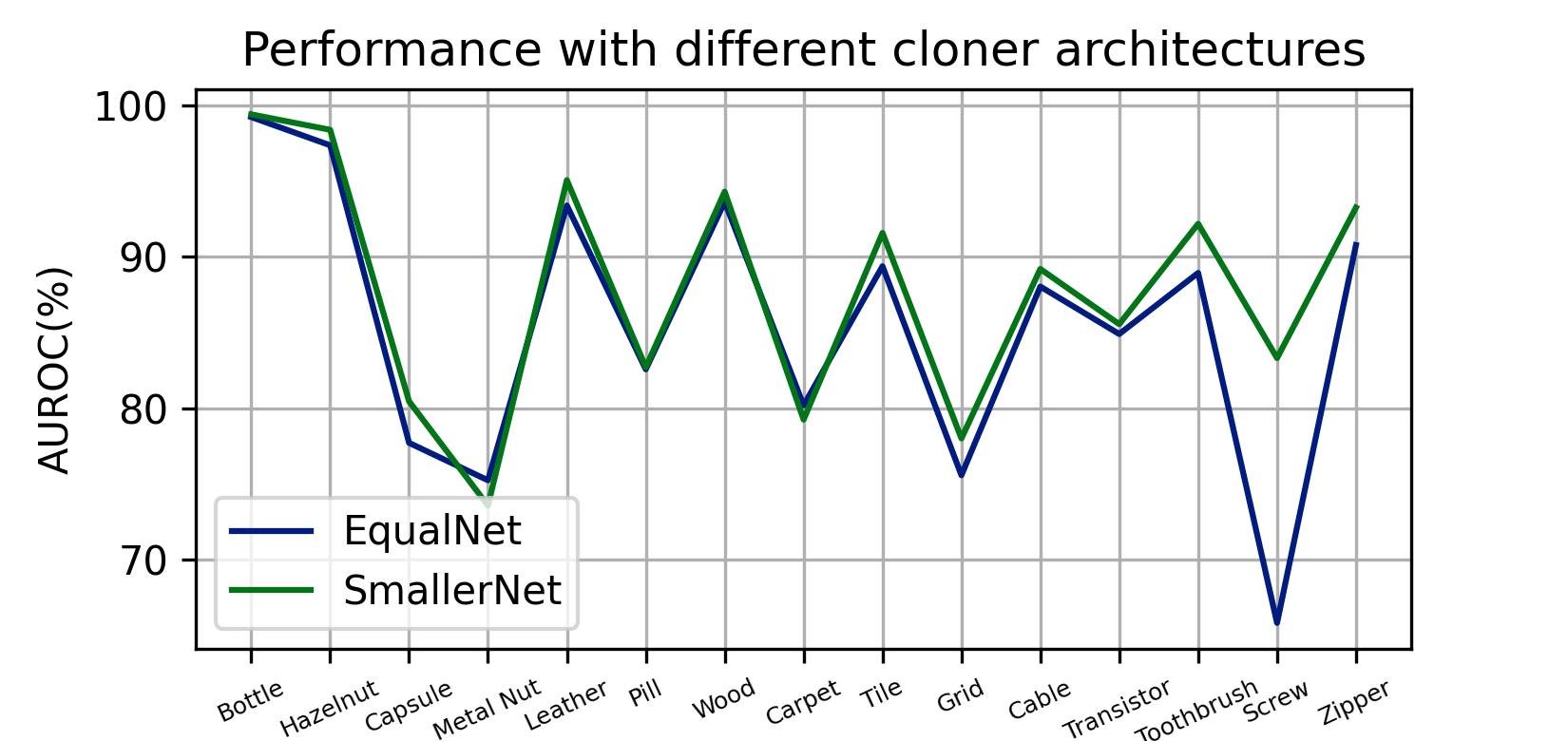}
      \caption{The performance of our proposed method using different equal/smaller cloner architectures compared to the source. 
      Smaller network performs better in general.}
       \label{fig:ablation_compression}
\end{figure}

As originally motivated in the knowledge distillation field, smaller $C$ plays an important role in our approach by eliminating non-distinguishing filters causing various distractions. It is especially more important when performing on normal data where the scope is dramatically limited. 
Here, we probe the effect of the cloner's architecture. As in Figure \ref{fig:ablation_compression}, anomaly detection, on MVTecAD, using a compact $C$ network outperforms a $C$ network with equal size to $S$. This is especially noticeable on classes in which anomalies are partial (like in ``toothbrush'' or ``screw''). Overall, the smaller network performs better with a margin of $\sim3\%$.

\subsubsection{$\mathcal{L}_{dir}$ and $\mathcal{L}_{val}$}
\label{ablation:dir_loss}
\begin{figure}[h]
\centering
     \includegraphics[width=\linewidth]{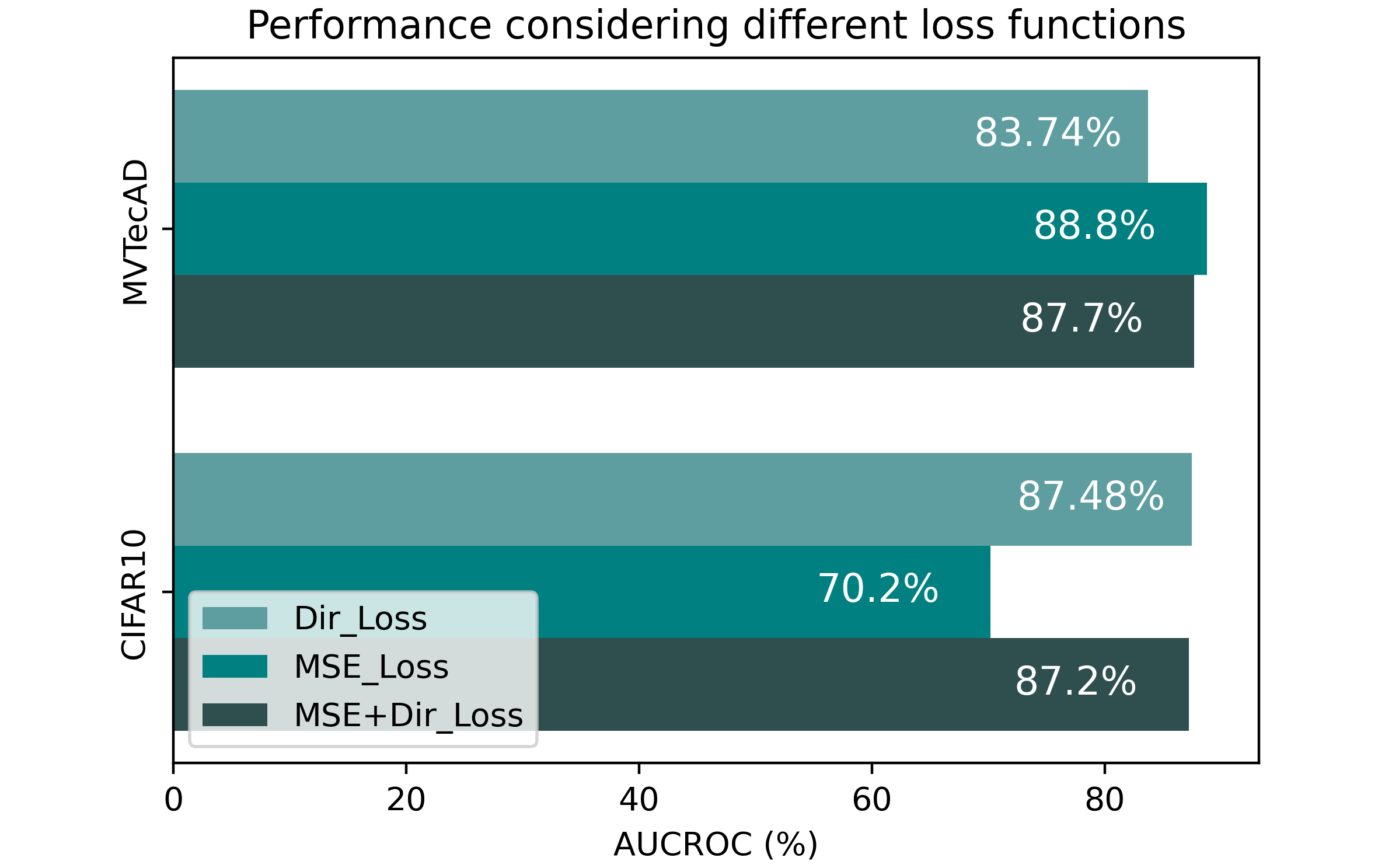}
      \caption{The performance of our proposed method using different loss functions. $\mathcal{L}_{total}$ performs well on both cases while individual directional or Euclidean losses fail in one.}
       \label{fig:ablation_loss}
\end{figure}

In this part, we discuss each loss component's effect to show the insufficiency of solely considering the Euclidean distance or directional loss in practice. The high impact of using $\mathcal{L}_{total}$ can be seen in Fig. \ref{fig:ablation_loss}. We report the mean AUROC over all the classes in the datasets. For more ablation studies, refer to Supplementary Materials for a a class-detailed report. Discarding the directional loss term drastically harms the overall performance on cases where anomalies are essentially different from normal cases and are more diverse, like in CIFAR-10. Using $\mathcal{L}_{dir}$ alone, however, shows top results. On the other hand, when considering cases with subtle anomalies MSE loss performs noticeably better and $\mathcal{L}_{dir}$ fails in comparison. However, in both cases, our proposed $\mathcal{L}_{total}$, which is a combination of the two losses, can achieve the highest performance. Theses results highlight the positive impact of considering a direction-wise notion of knowledge in addition to an MSE approach. 

\subsubsection{Localization using Interpretability Methods}
\label{ablation:localization}
\begin{table}[b]
    \centering
    \caption{Pixel-wise (AUROC) of anomaly localization on MVTecAD using different interpretability methods with and without Gaussian filtering.}
    \label{tab:ablation_local}
    \begin{tabular}{c|c|c|c}
         Method & Gradients & SmoothGrad & GBP\\
         \hline
         \hline
         \shortstack{Without \\ Gaussian Filter} & 86.16\% & 86.97\% & 84.38\%\\
         \hline
         \shortstack{With \\ Gaussian Filter} & 90.51\% & 90.54\% & 90.08\%\\
    \end{tabular}
\end{table}
In addition to simple gradients explained in Eq. \ref{local_eq}, in this section, other interpretability methods are also used for anomaly localization in our framework. In Table \ref{tab:ablation_local}, the results on MVTecAD images are shown with and without applying Guassian filter. As expected, SmoothGrad highlights the anomalous parts better than others as it discards wrongly highlighted pixels by Gradients, through calculating an average over gradients of noisy inputs. GBP, however, performs weaker than others since it tends more to reconstruct the image instead of staying faithful to the function \cite{adebayo2018sanity, nie2018theoretical}. Anyway, after applying the noise-removing filters, the methods perform almost the same. Hence, we use simple Gradients in the rest of our experiments instead of SmoothGrad that requires severe additional computations. 




\section{Experiments}

In this section, extensive  experiments have been done to demonstrate the effectiveness of our method.  \footnote{Code to reproduce the results is provided at \url{https://github.com/Niousha12/Knowledge_Distillation_AD}.}
Unlike other methods that report their maximum achieved results, we report an average on our trained models, sampled every 10 epochs after convergence, to show our training stability. Variances are also reported. Finally, we emphasize that $S$ is pre-trained on ImageNet and has not seen any data of the tested datasets. Hence, the comparison is fair.

\subsection{Experimental Setup}
\textbf{Datasets:}  We test our method on 7 datasets as follows:
\textbf{MNIST \cite{lecun2010mnist}:} 60k training and 10k test $28 \times 28$ gray-scale handwritten digit images.
\textbf{Fashion-MNIST \cite{xiao2017fashion}:} similar to MNIST (with 10k more training images) made up of 10 fashion product categories.
\textbf{CIFAR-10 \cite{cifar}} 50k training and 10k test $32 \times 32$ color images in 10 equally-sized natural entity classes.
\textbf{MVTecAD \cite{bergmann2019mvtec}:} an industrial dataset with over 5k high-resolution images in 15 categories of objects and textures. Each category has both normal images and anomalous images having various kinds of defects (only for testing). All images have been down scaled to the size $128 \times 128$.
\textbf{Retinal OCT Images (optical coherence tomography) \cite{gholami2020octid}:} consisting of 84,495 X-Ray images and 4 categories.
\textbf{HeadCT \cite{kitamura2018hemorrhage}:} a medical dataset containing 100 $128 \times 128$ normal head CT images and 100 with hemorrhage. Each image comes from a different person.
\textbf{BrainMRI for brain tumor detection \cite{chakrabarty2019tumor}:} consisting of 98 $256 \times 256$ normal MRI images and 155 with tumors.

\textbf{Evaluation Protocol:}
\label{sec:eval_protocol}
\textbf{Medical datasets:} 10 random normal images + all anomalous ones for test, the rest normal ones for training. \textbf{MVTecAD \& Retinal-OCT}: datasets train and test sets are used. \textbf{Others:} one class as normal and others as anomaly, at testing: the whole test set is used. 

\begin{figure*}[!t]
\centering
     \includegraphics[width=\textwidth]{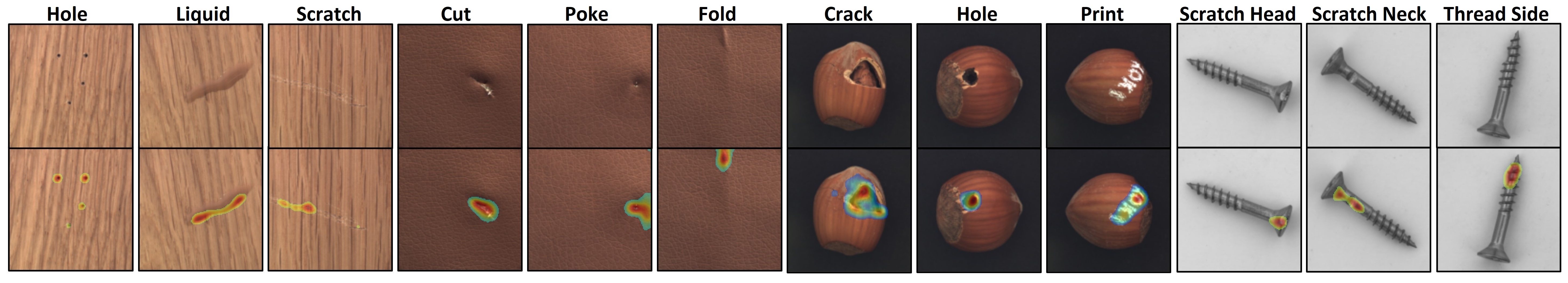}
      \caption{Anomaly localization map on different types of anomalies in MVTecAD dataset's sample classes. Pixels with low score are omitted from heatmap. This indicates our method's precise maps, no matter the defections' variety.}
       \label{fig:mvtec_localization}
\end{figure*}

\subsection{Results}
\subsubsection{MNIST \& Fashion-MNIST \& CIFAR10}

\begin{table*}[ht]
\centering
\caption{AUROC in \% for anomaly \textbf{detection} on \textbf{several} datasets. As shown, our model shows SOTA results on MNIST~\cite{lecun2010mnist} and Fashion-MNIST~\cite{xiao2017fashion}. On CIFAR-10~\cite{cifar} dataset our result is 13\% more than SOTA.}
\label{table:mnist,fashionmnist,CIFAR-10}
\resizebox{\textwidth}{!}{\begin{tabular}{ccccccccccccc}
\hline\noalign{\smallskip}
Dataset & Method & 0 & 1 & 2 & 3 & 4 & 5 & 6 & 7 & 8 & 9 & Mean\\
\noalign{\smallskip}
\hline
\noalign{\smallskip}
\multirow{9}{*}{MNIST\cite{lecun2010mnist}} & ARAE\cite{salehi2020arae} & 99.8 & 99.9 & 96.0 & 97.2 & 97.0 & 97.4 & 99.5 & 96.9 & 92.4 & 98.5 & 97.5\\
& OCSVM\cite{chen2001one} & 99.5 & 99.9 & 92.6 & 93.6 & 96.7 & 95.5 & 98.7 & 96.6 & 90.3 & 96.2 & 96.0\\
& AnoGAN\cite{schlegl2017unsupervised} & 96.6 & 99.2 & 85.0 & 88.7 & 89.4 & 88.3 & 94.7 & 93.5 & 84.9 & 92.4 & 91.3\\
& DSVDD\cite{ruff2018deep} & 98.0 & 99.7 & 91.7 & 91.9 & 94.9 & 88.5 & 98.3 & 94.6 & 93.9 & 96.5 & 94.8\\
& CapsNet\textsubscript{PP} \cite{li2019icml} & 99.8 & 99.0 & 98.4 & 97.6 & 93.5 & 97.0 & 94.2 & 98.7 & 99.3 & 99.0 & 97.7\\
& OCGAN\cite{perera2019ocgan} & 99.8 & 99.9 & 94.2 & 96.3 & 97.5 & 98.0 & 99.1 & 98.1 & 93.9 & 98.1 & 97.5\\
& LSA\cite{abati2019latent} & 99.3 & 99.9 & 95.9 & 96.6 & 95.6 & 96.4 & 99.4 & 98.0 & 95.3 & 98.1 & 97.5\\
& CAVGA-D\textsubscript{u}\cite{Venkataramanan2019attention} & 99.4 & 99.7 & 98.9 & 98.3 & 97.7 & 96.8 & 98.8 & 98.6 & 98.8 & 99.1 & 98.6\\
& U-Std\cite{Bergmann2020ustd} & 99.9 & 99.9 & 99 & 99.3 & 99.2 & 99.3 & 99.7 & 99.5 & 98.6 & 99.1 & \textbf{99.35}\\

\noalign{\smallskip}
\hline
\noalign{\smallskip}
& OURS & ${99.82\pm0.023}$ & ${99.82\pm0.017}$ & ${97.79\pm0.272}$ & ${98.75\pm0.098}$ & ${98.43\pm0.096}$ & ${98.16\pm0.182}$ & ${99.43\pm0.038}$ & ${98.38\pm0.178}$ & ${98.41\pm0.157}$ & ${98.1\pm0.152}$ & 98.71\\
\noalign{\smallskip}
\hline
\noalign{\smallskip}

\multirow{7}{*}{Fashion-MNIST\cite{xiao2017fashion}} & ARAE\cite{salehi2020arae} & 93.7 & 99.1 & 91.1 & 94.4 & 92.3 & 91.4 & 83.6 & 98.9 & 93.9 & 97.9 & 93.6\\
& OCSVM\cite{chen2001one} & 91.9 & 99.0 & 89.4 & 94.2 & 90.7 & 91.8 & 83.4 & 98.8 & 90.3 & 98.2 & 92.8\\
& DAGMM\cite{zong2018deep} & 30.3 & 31.1 & 47.5 & 48.1 & 49.9 & 41.3 & 42.0 & 37.4 & 51.8 & 37.8 & 41.7\\
& DSEBM\cite{zhai2016deep} & 89.1 & 56.0 & 86.1 & 90.3 & 88.4 & 85.9 & 78.2 & 98.1 & 86.5 & 96.7 & 85.5\\
& DSVDD\cite{ruff2018deep} & 98.2 & 90.3 & 90.7 & 94.2 & 89.4 & 91.8 & 83.4 & 98.8 & 91.9 & 99.0 & 92.8\\
& LSA\cite{abati2019latent} & 91.6 & 98.3 & 87.8 & 92.3 & 89.7 & 90.7 & 84.1 & 97.7 & 91.0 & 98.4 & 92.2\\
\noalign{\smallskip}
\hline
\noalign{\smallskip}
& OURS & ${92.5\pm0.298}$ & ${99.21\pm0.064}$ & ${92.48\pm0.255}$ & ${93.8\pm0.095}$ & ${92.95\pm0.159}$ & ${98.21\pm0.157}$ & ${84.87\pm0.126}$ & ${99.02\pm0.331}$ & ${94.33\pm0.164}$ & ${97.51\pm0.055}$ & \textbf{94.49}\\

\noalign{\smallskip}
\hline
\noalign{\smallskip}

\multirow{11}{*}{CIFAR-10\cite{cifar}} & ARAE\cite{salehi2020arae} & 72.2 & 43.1 & 69.0 & 55.0 & 75.2 & 54.7 & 70.1 & 51.0 & 72.2 & 40.0 & 60.23\\
& OCSVM\cite{chen2001one} & 63.0 & 44.0 & 64.9 & 48.7 & 73.5 & 50.0 & 72.5 & 53.3 & 64.9 & 50.8 & 58.56\\
& AnoGAN\cite{schlegl2017unsupervised} & 67.1 & 54.7 & 52.9 & 54.5 & 65.1 & 60.3 & 58.5 & 62.5 & 75.8 & 66.5 & 61.79\\
& DSVDD\cite{ruff2018deep} & 61.7 & 65.9 & 50.8 & 59.1 & 60.9 & 65.7 & 67.7 & 67.3 & 75.9 & 73.1 & 64.81\\
& CapsNet\textsubscript{PP}\cite{li2019icml} & 62.2 & 45.5 & 67.1 & 67.5 & 68.3 & 63.5 & 72.7 & 67.3 & 71.0 & 46.6 & 61.2\\
& OCGAN\cite{perera2019ocgan} & 75.7 & 53.1 & 64.0 & 62.0 & 72.3 & 62.0 & 72.3 & 57.5 & 82.0 & 55.4 & 65.66\\
& LSA\cite{abati2019latent} & 73.5 & 58.0 & 69.0 & 54.2 & 76.1 & 54.6 & 75.1 & 53.5 & 71.7 & 54.8 & 64.1\\
& DROCC\cite{Goyal2020} & 81.66 & 76.74 & 66.66 & 67.13 & 73.62 & 74.43 & 74.43 & 71.39 & 80.02 & 76.21 & 74.23\\
& CAVGA-D\textsubscript{u}\cite{Venkataramanan2019attention} & 65.3 & 78.4 & 76.1 & 74.7 & 77.5 & 55.2 & 81.3 & 74.5 & 80.1 & 74.1 & 73.7\\
& GT\cite{golan2018deep} & 76.2 & 84.8 & 77.1 & 73.2 & 82.8 & 84.8 & 82 & 88.7 & 89.5 & 83.4 & 82.3\\
& U-Std\cite{Bergmann2020ustd} & 78.9 & 84.9 & 73.4 & 74.8 & 85.1 & 79.3 & 89.2 & 83 & 86.2 & 84.8 & 81.96\\
\noalign{\smallskip}
\hline
\noalign{\smallskip}
& OURS & ${90.53\pm0.158}$ & ${90.35\pm0.797}$ & ${79.66\pm0.415}$ & ${77.02\pm0.51}$ & ${86.71\pm0.346}$ & ${91.4\pm0.279}$ & ${88.98\pm0.2}$ & ${86.78\pm0.595}$ & ${91.45\pm0.148}$ & ${88.91\pm0.349}$ & \textbf{87.18}\\
\noalign{\smallskip}
\hline
\noalign{\smallskip}
\end{tabular}}
\end{table*}
First, we evaluate our method on the conventional AD task on MNIST, Fashion-MNIST, and CIFAR-10 as described in Sec. \ref{sec:eval_protocol}. This targets detecting anomalies disparate from the normal samples in essence and not only slightly. 
As CFIAR-10 images are natural images, they have been resized and normalized according to ImageNet's properties. No normalization and resizing is done for other datasets.

For evaluation, similar to previous works, the area under the receiver operating characteristic curve (AUROC) is used. This allows comparison using different thresholds on the anomaly score. We compare our method with an exhaustive set of state-of-the-art approaches, including generative, self-supervised and autoencoder-based methods, in Table \ref{table:mnist,fashionmnist,CIFAR-10}. We outperform all other methods on F-MNIST and CIFAR-10, while staying comparatively well on MNIST, though avoiding complicated training procedures. Note that some methods, like U-Std, apply dataset-dependent fine-tuning.
We, however, avoid such fine-tunings.

\subsubsection{MVTecAD}
\begin{table*}[!t]
\centering
\caption{AUROC in \% for anomaly \textbf{detection} on MVTecAD~\cite{bergmann2019mvtec}. We surpass the SOTA by $\sim10\%$}
\label{table:MVTec}
\resizebox{\textwidth}{!}{\begin{tabular}{ccccccccccccccccc}
\hline\noalign{\smallskip} 
Method & Bottle & Hazelnut & Capsule & Metal Nut & Leather & Pill & Wood & Carpet & Tile & Grid & Cable & Transistor & Toothbrush & Screw & Zipper & Mean\\
\noalign{\smallskip}
\hline
\noalign{\smallskip}
AVID \cite{sabokrou2018avid} & 88 & 86 & 85 & 63 & 58 & 86 & 83 & 70 & 66 & 59 & 64 & 58 & 73 & 66 & 84 & 73\\
AE\textsubscript{SSIM} \cite{bergmann2019visigrapp} & 88 & 54 & 61 & 54 & 46 & 60 & 83 & 67 & 52 & 69 & 61 & 52 & 74 & 51 & 80 & 63\\
AE\textsubscript{L2} \cite{bergmann2019visigrapp} & 80 & 88 & 62 & 73 & 44 & 62 & 74 & 50 & 77 & 78 & 56 & 71 & 98 & 69 & 80 & 71\\
AnoGAN\cite{schlegl2017unsupervised} & 69 & 50 & 58 & 50 & 52 & 62 & 68 & 49 & 51 & 51 & 53 & 67 & 57 & 35.0 & 59 & 55\\
LSA\cite{abati2019latent} & 86 & 80 & 71 & 67 & 70 & 85 & 75 & 74 & 70 & 54 & 61 & 50 & 89 & 75 & 88 & 73\\
CAVGA-D\textsubscript{U}\cite{Venkataramanan2019attention} & 89 & 84 & 83 & 67 & 71 & 88 & 85 & 73 & 70 & 75 & 63 & 73 & 91 & 77 & 87 & 78\\
DSVDD\cite{ruff2018deep} & 86 & 71 & 69 & 75 & 73 & 77 & 87 & 54 & 81 & 59 & 71 & 65 & 70 & 64 & 74 & 72\\
VAE-grad\cite{Dehaene2020} & 86 & 74 & 86 & 78 & 71 & 80 & 89 & 67 & 81 & 83 & 56 & 70 & 89 & 71 & 67 & 77\\
GT\textsuperscript{*}\cite{golan2018deep} & 74.29 & 33.32 & 67.79 & 82.37 & 82.51 & 65.16 & 48.24 & 45.9 & 53.86 & 61.91 & 84.7 & 79.79 & 94 & 44.58 & 87.44 & 67.06\\
\noalign{\smallskip}
\hline
\noalign{\smallskip}

OURS & ${99.39\pm0.032}$ & ${98.37\pm0.285}$ & ${80.46\pm0.19}$ & ${73.58\pm0.376}$ & ${95.05\pm0.208}$ & ${82.7\pm0.646}$ & ${94.29\pm0.226}$ & ${79.25\pm0.8}$ & ${91.57\pm0.66}$ & ${78.01\pm0.621}$ & ${89.19\pm0.378}$ & ${85.55\pm0.212}$ & ${92.17\pm0.323}$ & ${83.31\pm0.643}$ & ${93.24\pm0.247}$ & \textbf{87.74}\\
\noalign{\smallskip}
\hline
\noalign{\smallskip}
\end{tabular}}
\end{table*}
\textbf{Detection:} In this part, we report the results of our method performance on AD using MVTecAD. As shown in Table \ref{table:MVTec}, our method outperforms all others with a large margin of $\sim10\%$. This is remarkable since other methods fail to perform well in both one-class setting and defect detection simultaneously. In contrast, we achieve SOTA in both cases. 
\begin{table*}[!t]
\centering
\caption{AUROC in \% for anomaly \textbf{localization} on MVTecAD~\cite{bergmann2019mvtec}.}
\label{table:MVTec-local}
\resizebox{\textwidth}{!}{\begin{tabular}{ccccccccccccccccc}
\hline\noalign{\smallskip} 
Method & Bottle & Hazelnut & Capsule & Metal Nut & Leather & Pill & Wood & Carpet & Tile & Grid & Cable & Transistor & Toothbrush & Screw & Zipper & Mean\\
\noalign{\smallskip}
\hline
\noalign{\smallskip}
AE\textsubscript{SSIM}\cite{bergmann2019visigrapp} & 93 & 97 & 94 & 89 & 78 & 91 & 73 & 87 & 59 & 94 & 82 & 90 & 92 & 96 & 88 & 87\\
AE\textsubscript{L2}\cite{bergmann2019visigrapp} & 86 & 95 & 88 & 86 & 75 & 85 & 73 & 59 & 51 & 90 & 86 & 86 & 93 & 96 & 77 & 82\\
AnoGAN\cite{schlegl2017unsupervised} & 86 & 87 & 84 & 76 & 64 & 87 & 62 & 54 & 50 & 58 & 78 & 80 & 90 & 80 & 78 & 74\\
CNN-Dict\cite{Napoletano2018cnn-based}& 78 & 72 & 84 & 82 & 87 & 68 & 91 & 72 & 93 & 59 & 79 & 66 & 77 & 87 & 76 & 78\\
VAE-grad\cite{Dehaene2020} & 92.2 & 97.6 & 91.7 & 90.7 & 92.5 & 93 & 83.8 & 73.5 & 65.4 & 96.1 & 91 & 91.9 & 98.5 & 94.5 & 86.9 & 89.3\\ 
\noalign{\smallskip}
\hline
\noalign{\smallskip}
OURS & 96.32 & 94.62 & 95.86 & 86.38 & 98.05 & 89.63 & 84.8 & 95.64 & 82.77 & 91.78 & 82.4 & 76.45 & 96.12 & 95.96 & 93.9 & \textbf{90.71}\\
\noalign{\smallskip}
\hline
\noalign{\smallskip}
\end{tabular}}
\end{table*}
\begin{table*}[h!]
\centering
\caption{AUROC in \% on Retinal-OCT~\cite{Zhou2020Retina}. We outperform all other SOTA methods.}
\label{table:Retina}
\resizebox{\textwidth}{!}{\begin{tabular}{cccccccccccc}
\hline\noalign{\smallskip} 
& DSVDD\cite{ruff2018deep} & Auto-Encoder\cite{Zhou2017autoencoder} & AnoGan\cite{schlegl2017unsupervised} & VAE-GAN\cite{Baur2018vae-gan} & Pix2Pix\cite{Isola2017pix2pix} & GANomaly\cite{akcay2018ganomaly} & Cycle-GAN\cite{Zhu2017cycle-gan} & P-Net\cite{Zhou2020Retina} & GT\cite{golan2018deep} & OURS\\
\noalign{\smallskip}
\hline
\noalign{\smallskip}
RESC (OCT)\cite{gholami2020octid}
& 74.40 & 82.07 & 84.81 & 90.64 & 79.34 & 91.96 & 87.39 & 92.88 & 60.13 & ${\textbf{97.01}\pm\textbf{0.426}}$\\
\noalign{\smallskip}
\hline
\noalign{\smallskip}
\end{tabular}}
\end{table*}
\textbf{Localization:} We not only accomplish SOTA in AD but outperform previous SOTA methods in anomaly localization. As stated in \ref{ablation:localization}, we use simple gradients to obtain maps.
We use Gaussian filter with $\sigma=4$ and a $3 \times 3$ ellipse structuring element kernel. We compare our method against others, including AE-based and generative methods in Table \ref{table:MVTec-local}. We use AUROC, based on each pixel anomaly score, to measure how well anomalies are localized. Vividly, we outperform all previous methods. Fig. \ref{fig:mvtec_localization} shows our localization maps on different defects' types in MVTecAD.
\subsubsection{Medical Datasets}
To further evaluate our method in various domains, we use 3 medical datasets and compare ours method on them against others. First, we use Retinal-OCT dataset, a recent dataset for detecting abnormalities in retinal optical coherence tomography (OCT) images. According to Table \ref{table:Retina}, our method outplays all SOTA methods by a huge margin. This shows that the knowledge of the pre-trained netowrk, $S$, has been highly valuable to the cloner, $C$, even in an entirely different domain of medical retinal OCT inputs. Furthermore, the unawareness of $C$ about the outside of the normal data manifold, in contrast to $S$, intensifies the discrepancy between them. This expresses the generality of our method to even future unseen datasets, something missed in many methods. 

Moreover, we validate our performance on brain tumor detection using brain MRI images. In this dataset, images with tumors are assumed as anomalous while healthy ones are considered as normal. In Table \ref{table:Medical_Tumor}, our method achieves SOTA results alongside LSA. While slightly ($\sim 0.5\%$) less than LSA, our method shows a significantly less variance, magnifying its stability, compared to others. It is also noteworthy that LSA fails substantially on other tasks such as on CIFAR10 and MVTecAD anomaly detection with AUROCs $\sim 23\%$ and $\sim 25\%$ below our method's, respectively.

\begin{table}[h!]
\centering
\caption{AUROC in \%  medical datasets. The top two methods are in bold.}
\label{table:Medical_Tumor}
{\begin{tabular}{ccc}
\hline\noalign{\smallskip} 
& BrainMRI & HeadCT \\
\noalign{\smallskip}
\hline
\noalign{\smallskip}
LSA\textsuperscript{*}\cite{abati2019latent} & ${\textbf{95.61}\pm\textbf{1.433}}$ & ${\textbf{81.67}\pm\textbf{0.358}}$ \\
OCGAN\textsuperscript{*}\cite{perera2019ocgan} & 
${91.74\pm3.050}$ & ${51.22\pm3.626}$ \\
GT\textsuperscript{*}\cite{golan2018deep} & 
${80.82\pm1.996}$ & ${49.85\pm3.873}$ \\
\noalign{\smallskip}
\hline
\noalign{\smallskip}
OURS & ${\textbf{95.01}\pm\textbf{0.229}}$ & ${78.04\pm0.225}$ \\
OURS+AUG & - & ${\textbf{80.42}\pm\textbf{0.006}}$ \\
\noalign{\smallskip}
\hline
\noalign{\smallskip}
\end{tabular}}
\end{table}

Lastly, using HeadCT (hemorrhage) dataset, we discuss an important aspect of our model. Performing on head computed tomography (CT) images for AD, we ouperform OCGAN and GT by a huge margin, and perform $\sim 3\%$ below LSA. Here, since the training data is dramatically limited, our method can possibly face difficulties transferring the $S$'s knowledge to $C$. However, this can be addressed by using simple data augmentations. We use 20 degree rotation in addition to scaling in range $[ 0.9, 1.05]$ to augment the images. These augmentations are generic non-tuned ones aiming solely to increase the amount of data with no dependency to the dataset. In Table \ref{table:Medical_Tumor}, it is showed that using augmentation, the proposed method achieves similar results to LSA's, while outshining it on other tasks significantly.

\section{Conclusion}
We show that ``distilling" the intermediate knowledge of an ImageNet pre-trained expert network on anomaly-free data into a more compact cloner network, and then using their different behavior with different samples, sets a new direction for finding distinctive criterion to detect and localize anomalies. Without using intensive region-based training and testing, we leverage interpretability methods in our novel framework for obtaining localization maps. We achieve superior results in various tasks and on many datasets even with domains far from ImageNet's domain. 

{\small
\bibliographystyle{ieee_fullname}

}

\twocolumn[{%
 \centering
 \LARGE Appendix\\[1em]
}]

\appendix
We report the results of the ablation studies in the paper on different classes of MVTecAD, CIFAR10, and MNIST in more details here.
\section{Intermediate Knowledge}
The performance of our framework using different layers as critical points for distillation was discussed in Sec. 3.3.1. Here, we provide the class-detailed performance on MVTecAD and MNIST in Table~\ref{table:MVTec-supp-layers} and Table~\ref{table:mnist-supp-layers}. As discussed in the paper, the performance is enhanced when more intermediate hints are considered. Note that the ``only the last layer'' setting performs roughly the same as a random detector (AUC=50\%) on some MVTecAD classes.
\begin{table*}[p]
\centering
\caption{Class-detailed AUROC of our proposed method using various layers for distillation. More intermediate layers lead to a performance boost on anomaly detection on MVTecAD.}
\label{table:MVTec-supp-layers}
\resizebox{\textwidth}{!}{\begin{tabular}{cccccccccccccccc|c}
& Bottle & Cable & Capsule & Carpet & Grid & Hazelnut & Leather & Metal nut & Pill & Screw & Tile & Toothbrush & Transistor & Wood & Zipper & \textbf{Mean}\\
\hline
\hline
TheLast & 99.6 & 82.6 & 79.4 & 72.8 & 48.9 & 91.0 & 83.6 & 76.1 & 66.2 & 59.4 & 82.6 & 85.5 & 87.6 & 83.9 & 89.1 & 79.22 \\
\hline
TheLast2 & 99.2 & 89.19 & 76.8 & 74.1 & 58.2 & 96.3 & 86.6 & 78.1 & 75.3 & 72.2 & 86.5 & 83.4 & 85.2 & 95.4 & 89.8 & 83.02 \\
\hline
TheLast4 & 99.4 & 98.4 & 80.5 & 73.6 & 95.1 & 82.7 & 94.3 & 79.3 & 91.6 & 78.0 & 89.2 & 85.6 & 92.2 & 83.3 & 93.2 & \textbf{87.74}\\
\hline
\noalign{\smallskip}

\end{tabular}}
\end{table*}
\begin{table*}[p]
\centering
\caption{Class-detailed AUROC of our proposed method using various layers for distillation. More intermediate layers lead to a performance boost on anomaly detection on MNIST.}
\label{table:mnist-supp-layers}

{\begin{tabular}{ccccccccccc|c}
& 0 & 1 & 2 & 3 & 4 & 5 & 6 & 7 & 8 & 9 & \textbf{Mean}\\ \hline
\hline
TheLast & 97.65 & 98.87 & 93.27 & 95.10 & 95.19 & 94.95 & 97.63 & 93.14 & 94.62 & 92.93 & 95.33\\
\hline
TheLast2 & 99.39 & 99.60 & 96.80 & 97.68 & 97.94 & 97.10 & 98.85 & 96.86 & 96.6 & 96.62 & 97.74\\
\hline
TheLast3 & 99.82 & 99.82 & 97.79 & 98.75 & 98.4 & 98.16 & 99.43 & 98.38 & 98.41 & 98.1 & \textbf{98.71}\\
\hline
\noalign{\smallskip}

\end{tabular}}
\end{table*}

\section{Distillation Effect (Compact Cloner)}
In this section, we provide the details of results in Sec. 3.3.2 of the paper. As mentioned in the paper, using a more compact cloner network outperforms when a network with equal size to the source is employed for cloner. Here in Tables. \ref{table:MVTec-Distilation-loss-Detail} and \ref{table:Cifar-Distilation-loss-Detail} we present a class-detailed comparison for MVTecAD and CIFAR-10 datasets.
\begin{table*}[p]
\centering
\caption{The detailed AUROC of our method using different loss functions and equal/smaller cloner architectures compared to the source. Both reported on MVTecAD classes. $\mathcal{L}_{total}$ performs well on both cases while individual directional or Euclidean losses fail in one. Also, smaller network for the cloner performs better in general.}
\label{table:MVTec-Distilation-loss-Detail}
\resizebox{\textwidth}{!}{\begin{tabular}{cccccccccccccccc|c}
& Bottle & Cable & Capsule & Carpet & Grid & Hazelnut & Leather & Metal nut & Pill & Screw & Tile & Toothbrush & Transistor & Wood & Zipper & \textbf{Mean}\\
\hline
EqualNet & 99.2 & 88.0 & 77.7 & 80.2 & 75.6 & 97.4 & 93.4 & 76.3 & 82.6 & 65.8 & 89.4 & 88.9 & 84.9 & 93.6 & 90.8 & 85.58 \\
SmallerNet & 99.4 & 89.2 & 80.5 & 79.3 & 78.0 & 98.4 & 95.1 & 73.6 & 82.7 & 83.3 & 91.6 & 92.2 & 85.6 & 94.3 & 93.2 & \textbf{87.74} \\
\hline
Dir Loss & 99.4 & 89.3 & 78.8 & 74.1 & 50.1 & 98.1 & 92.3 & 81.6 & 77.8 & 63.5 & 91.2 & 92.0 & 87.7 & 88.1 & 92.2 & 83.74 \\
MSE Loss & 99.4 & 87.6 & 81.3 & 81.3 & 82.3 & 98.1 & 94.3 & 71.6 & 85.3 & 92.3 & 90.7 & 94.0 & 83.9 & 96.0 & 94.6 & 88.8 \\
Total Loss & 99.4 & 89.2 & 80.5 & 79.3 & 78.0 & 98.4 & 95.0 & 73.6 & 82.7 & 83.3 & 91.6 & 92.2 & 85.6 & 94.3 & 93.2 & 87.74 \\
\hline
\noalign{\smallskip}

\end{tabular}}
\end{table*}
\begin{table*}[p]
\centering
\caption{The detailed AUROC of our method using different loss functions and equal/smaller cloner architectures compared to the source. Both reported on CIFAR-10 classes. $\mathcal{L}_{total}$ performs well on both cases while individual directional or Euclidean losses fail in one. Also, smaller network for the cloner performs better in general.}
\label{table:Cifar-Distilation-loss-Detail}

{\begin{tabular}{ccccccccccc|c}
& Airplane & Car & Bird & Cat & Deer & Dog & Frog & Horse & Ship & Truck & \textbf{Mean} \\
\hline
Equal Net & 90.04 & 89.89 & 80.97 & 77.24 & 86.88 & 91.38 & 87.72 & 84.48 & 90.80 & 89.34 & 86.87 \\
Smaller Net & 90.53 & 90.35 & 79.66 & 77.02 & 86.71 & 91.40 & 88.98 & 86.78 & 91.45 & 88.91 & 87.18 \\
\hline
Dir Loss & 90.42 & 91.07 & 79.41 & 76.98 & 86.69 & 91.72 & 89.21 & 87.69 & 91.36 & 90.27 & 87.48 \\
MSE Loss & 77.36 & 61.96 & 66.75 & 58.94 & 83.21 & 60.38 & 81.67 & 67.17 & 79.29 & 64.88 & 70.16 \\
Total Loss & 90.53 & 90.35 & 79.66 & 77.02 & 86.71 & 91.40 & 88.98 & 86.78 & 91.45 & 88.91 & 87.2 \\
\hline
\noalign{\smallskip}

\end{tabular}}
\end{table*}


\section{$\mathcal{L}_{dir}$ and $\mathcal{L}_{val}$}
In this part, we present a classed-detailed report for the effect of each loss component as discussed in Sec. 3.3.3 in the paper. We report the AUROC for all the classes in MVTecAD and CIFAR-10 datasets in Table \ref{table:MVTec-Distilation-loss-Detail} and Table \ref{table:Cifar-Distilation-loss-Detail} respectively. As investigated in the paper, $\mathcal{L}_{total}$, which is a combination of the directional and MSE loss, achieves the best performance. Theses results highlight the positive impact of considering a direction-wise notion of activations' knowledge in addition to an MSE approach. 

\section{Localization using Interpretability Methods}
Here, we report detailed results of Sec. 3.3.4 in the paper. In Table. \ref{table:MVTec-ablation-algorithm}, the AUROC for all MVTecAD classes are shown with and without applying the Guassian filter. As discussed in the paper, SmoothGrad highlights the anomalous parts better than others and GBP performs weaker than others. Anyway, after applying the noise-removing filters, the methods perform almost the same.
\begin{table*}[!ht]
\centering
\caption{Pixel-wise (AUROC) of anomaly localization on
MVTecAD using different interpretability methods with
and without Gaussian filtering. Without applying the filters, SmoothGrad outperforms others. With Gaussian filtering the methods perform almost the same.}
\label{table:MVTec-ablation-algorithm}
\resizebox{\textwidth}{!}{\begin{tabular}{c|cccccccccccccc}
& Gradients + Gaussian & Gradients & SmoothGrad + Gaussian & SmoothGrad & GBP + Gaussian & GBP \\
\hline
\hline
Bottle & 96.32 & 93.2 & 96.03 & 93.91 & 95.08 & 90.46 \\
Cable & 82.4 & 76.24 & 85.64 & 81.3 & 80.21 & 72.34 \\
Capsule & 95.86 & 93.06 & 95.55 & 93.45 & 95.43 & 91.53 \\
Carpet & 95.64 & 90.97 & 95.48 & 92.98 & 94.95 & 90.2 \\
Grid & 91.78 & 84.07 & 91.4 & 86.44 & 90.44 & 81.46 \\
Hazelnut & 94.62 & 91.3 & 94.33 & 89.96 & 95.06 & 91.09 \\
Leather & 98.05 & 95.41 &  98.04 & 96.76 & 97.96 & 94.32 \\
Metal nut & 86.38 & 82.15 & 86.15 & 82.54 & 83.45 & 77.73 \\
Pill & 89.63 & 86.33 & 88.99 & 85.07 & 90.32 & 84.99 \\
Screw & 95.96 & 93.42 & 94.34 & 91.3 & 95.3 & 93.03 \\
Tile & 82.77 & 77.4 & 82.92 & 79.37 & 82.6 & 76.47 \\
Toothbrush & 96.12 & 92.13 & 95.64 & 92.14 & 95.3 & 90.28 \\
Transistor & 76.45 & 71.02 & 76.54 & 73.13 & 76.49 & 68.84 \\
Wood & 84.8 & 78.53 & 83.4 & 78.95 & 84.85 & 77.47 \\
Zipper & 93.9 & 87.23 & 93.64 & 87.18 & 93.81 & 85.51 \\
\hline
Mean & 90.71 & 86.16 & 90.54 & 86.97 & 90.08 & 84.38 \\
\hline
\noalign{\smallskip}

\end{tabular}}
\end{table*}

\end{document}